# Blind Stereo Image Quality Assessment Inspired by Brain Sensory-Motor Fusion

Maryam Karimi, Najmeh Soltanian, Shadrokh Samavi, Nader Karimi, S.M.Reza Soroushmehr, Kayvan Najarian

*Abstract*— **The use of 3D and stereo imaging is rapidly increasing. Compression, transmission, and processing could degrade the quality of stereo images. Quality assessment of such images is different than their 2D counterparts. Metrics that represent 3D perception by human visual system (HVS) are expected to assess stereoscopic quality more accurately. In this paper, inspired by brain sensory/motor fusion process, two stereo images are fused together. Then from every fused image two synthesized images are extracted. Effects of different distortions on statistical distributions of the synthesized images are shown. Based on the observed statistical changes, features are extracted from these synthesized images. These features can reveal type and severity of distortions. Then, a stacked neural network model is proposed, which learns the extracted features and accurately evaluates the quality of stereo images. This model is tested on 3D images of popular databases. Experimental results show the superiority of this method over state of the art stereo image quality assessment approaches.**

## I. INTRODUCTION

NOWADAYS, with the expansion of communication through internet and other communication networks, high volume of media is being transferred. The quality of delivered images needs to be assured using visual media quality assessment (QA). Although most sensible methods for determining the quality of images and video are subjective assessment, they are impractical due to being laborious, costly, and time-consuming. Moreover, subjective assessment is inefficient for real-time applications and always depends on physical, emotional and individual differences of people [1]. Therefore, a lot of research has been done to design an automated computational model for objective quality assessment of images [2-6]. Since any proposed model must estimate scores close to subjective ratings, successful objective methods have high correlated results with human observations.

The popularity of three-dimensional images and videos in recent years is increasing [7]. The number of three-dimensional movies increases at least 50% each year [8]. In addition to movies, 3D televisions and cameras have become commonplace. Scientific applications, medical and military usage of 3D images are not negligible. Therefore, it is expected that in the near future 3D media covers a large portion of all transferred data. Hence, monitoring and quality protection of visual content for such images is one of the new challenges ahead. So far, extensive research has been done to determine the quality of 2D images [2-6], but research in the field of 3D images is relatively recent [9-40]. Appropriate, efficient, and fast solutions for assessment of such images can help development of 3D imaging applications. With the addition of depth as the third dimension, new issues such as depth perception, visual discomfort, visual fatigue and visual perception arise that make 3D Stereo Image Quality Assessment (SIQA) much more complex than its 2D counterpart. These issues make 3D visual quality assessment very sensitive and quite challenging field of research. Therefore, using routines developed for 2D images is inappropriate for 3D images and new methods are required to address this problem [9].

The distortion topic in stereo images covers many details such as visual discomfort, unbalanced depth perception and visual fatigue due to incorrect stereography. The scenario of SIQA assesses the quality of stereo images that independently or simultaneously (symmetric or asymmetric) have been affected by distortions such as compression, noise or camera artifacts. Distorted images are well-calibrated, which means there is no problem in terms of camera angles and stereography settings. Similar to 2D images, 3D visual quality evaluation methods are divided into three following categories based on the need of the method to access the original (reference) image:

1) Full-reference (FR) models need the original image pair to assess the quality of degraded image pair.

2) Reduced-Reference (RR) methods have access to some features or some information about the original images.

3) Unlike the first two categories, No-Reference (NR) algorithms estimate visual quality of degraded stereo images without any information or any need to the reference images.

Maryam Karimi is with the Department of Electrical Engineering, Isfahan University of Technology, Isfahan 84156-83111, Iran.
Najmeh Soltanian is with the Department of Electrical Engineering, Isfahan University of Technology, Isfahan 84156-83111, Iran.
Shadrokh Samavi is with the Department of Electrical and Computer Engineering, Isfahan University of Technology, Isfahan 84156-83111, Iran, and McMaster University, Hamilton, Canada.
Nader Karimi is with the Department of ECE, Isfahan University of Technology, Iran.
S.M.Reza Soroushmehr is with the Dept. of Computational Medicine and Bioinformatics, University of Michigan, Ann Arbor, U.S.A.
Kayvan Najarian is with the Michigan Center for Integrative Research in Critical Care, and also with the Dept. of Computational Medicine and Bioinformatics, University of Michigan, Ann Arbor, U.S.A.

Since the original versions of received signals from communication channels are not available in most cases, NR IQA methods are more practical than the other two categories. In the case of stereo images, availability of reference images means having both the left and right images. The application of FR SIQA methods is more limited than the application of FR 2D methods. Therefore, NR and RR methods are the main priority of stereo quality assessment systems. Despite 2D NR QA methods which produce comparable results to 2D FR ones [4-6], 3D NR QA approaches are not as strong as 3D FR methods.

From another perspective, QA techniques can be divided into two categories of general purpose and application specific techniques. General purpose methods estimate the degree of image quality, independent of the type of distortion. These methods are flexible and are based on common characteristics and assumptions about human visual system. Application specific criteria are designed for a specific use.

In this paper, a general purpose no-reference stereo image quality assessment (NR SIQA) approach based on a stacked structure of an artificial neural network (ANN) is proposed. Our work is based on the fact that human brain generates a binocular combination of two images, called cyclopean image. Discomforts for the visual system occur by the distortions in the perceived depth or by distortions in the spatial domain of the cyclopean perception [9]. Hence, rather than analyzing the right and left images individually, we generate a cyclopean image based on motor/sensory fusion process of the brain. The synthesized cyclopean perception could be analyzed into "*phase*" and "*contrast*" matrices or images. To further imitate the human's brain behavior, we use neural networks to separately study each of the two extracted images. Then the outputs of these two networks are fed into another neural network to generate the final assessment of the stereo images. Our contributions can be summarized as: (a) use of binocular combined images with maximum coverage of the visual discomfort characteristics, (b) proper generation of features to reveal wide range of possible distortions, and severities of distortions, (c) proper use of stacked neural networks. These enable our system to outperform exiting SIQA algorithms.

The rest of this paper is organized as follows. In Section II, a review of objective stereo image quality assessment is presented. In Section III, we develop our model, describing in detail, the image combination, feature extraction and learning based quality estimation mechanisms. Section IV, describes the experimental results on related stereo image databases and Section V concludes the paper.

## II. RELATED WORK

The necessity of automatic media quality monitoring in recent years has attracted a lot of research in the field of objective quality assessment. The most recent challenge in this field is 3D image and video quality assessment. Visual quality evaluation of 3D images is a complex issue which is not easily understood, analyzed, and solved. In the following we review some quality assessment methods for 3D images which could be classified into three groups.

### A. Full Reference Methods

Most of the FR 3D quality assessment metrics either evaluate the quality of left and right images using 2D image quality assessment algorithms or evaluate the difference between test and reference depth maps. Authors of [10] have studied the use of two 2D FR metrics, Structural Similarity Index Measure (SSIM [2]), Universal Quality Index (UQI [11]) and a 2D RR metric in [12], to evaluate the quality of 3D images. In [13], numerous 2D metrics are employed to estimate quality of color plus depth encoded video. Authors of [14], in addition to disparity information, take advantage of 10 well known 2D FR metrics to determine the quality of the stereo images. In [15], the authors concluded that using only 2D metrics is not sufficient. They improved results with contour analysis of synthesized view and mean SSIM calculation in disoccluded regions. The method presented in [16] combines stereo similarity map and disparity map for 3D quality assessment. Benoit *et al.* improved SSIM for JPEG, JPEG2000 and blurred images using additional depth information [17]. In addition to left and right image qualities, the disparity quality of the distorted pair has been considered for quality assessment [18, 19]. However, the problem of these methods is that areas with low disparities are always considered as lower quality areas and the impact of differences in disparity map is assumed to be the same everywhere. To assess the quality of stereoscopic video in [20] the influence of different depth layers in image quality has been studied. The authors found that the quality of low disparity areas and video content types effect the 3D visual quality.

Another group of FR methods exploits the characteristics of binocular vision to assess the quality of stereo images. A metric introduced in [21] that obtains binocular energy of the left and right images regarding to spatial frequency in different orientations and channels. Based on amplitude changes of this energy, it estimates the depth quality that is reconstructed by HVS. In [22], similar blocks in the left and right images are analyzed by 3D-DCT where mean squared errors measured in 3D-DCT domain are used for estimation of contrast sensitivity and luminance masking characteristics in HVS. Ryu *et al.* proposed a stereo version of the SSIM based on binocular quality perception and combined luminance similarity, contrast similarity, and structural similarity with binocular quality perception model to form the final quality index [23]. The proposed metric in [24] is a three-stage model based on BJND. In this method after developing a perceptual representation for each image, BJND models for the reference and distorted images are formed by independent assessment of pixels in different classes. The final score is calculated by averaging all the assessments. The algorithm presented in [25], computes the quality score by applying Multi Scale SSIM (MS-SSIM) to a weighted sum of stereo images called combined cyclopean image. The weight values are based on the response of Gabor filter bank. In [26] an FR metric is introduced which uses binocular combination behavior to enhance the performance of SIQA models. It first produces two channels of summation and difference from the two reference input images and the two test images. Then it generates a weighted combination of these two channels and exposes this combination to FR quality metrics. In [27], by performing consistency check, the left and right images are divided into three areas. Then, each region is assessed based on amplitude and phase maps of the reference and distorted images independently. Finally, region scores are combined to achieve a quality score. Lin and Wu decomposed the reference pair and the degraded pair of images into different spatial/frequency ranges by employing Difference of Gaussian (DOG) filter bank [28]. The final quality score is a weighted sum of quality scores in different frequency ranges. The method [29] learns a multi scale dictionary from the training dataset. Then the difference of sparse coefficient vectors of reference and test images are used to compute the similarity index. The final quality score is the binocular combination of the left and right indices.

### B. Reduced-Reference and No-Reference Methods

Reduced Reference and NR image quality assessment methods provided for stereo images till now are very limited. In [30], the edge information of the reference depth map is sent as the reduced reference information. The overall quality is achieved from the PSNR of depth maps. The metric presented in [31] achieved by comparing sensitivity coefficients of cyclopean images as well as coherence between their disparity maps. The approaches of [32] are to consider color and depth as the sensation of depth in 3D video. In model [33], after a divisive normalization in contourlet domain for the image and disparity map, the feature parameters of the fitted Gaussian distributions are used to prepare the quality metric.

In [34], an NR metric in synthesized image domain based on temporal outliers, temporal inconsistencies, and spatial outliers is proposed. In [35], an NR quality score for encoded images and video were estimated using local information of distorted stereo image and its disparity map. The method presented in [36] measures luminance and contrast distortion in synthesized images. Sohn and Ryu proposed an NR method to assess the blurriness and blockiness in binocular vision while blurriness, blockiness, and saliency maps were extracted from left and right images [37]. By combining blurriness and blockiness scores of each image, the final score was calculated. Akhter *et al.* [38] suggested an NR quality assessment method for JPEG stereo images. In this method the blockiness and blurriness scores of the left and right images, in addition to features extracted from imprecise disparity, are combined with each other. Chen *et al.* [9] use Blind/Referenceless Image Spatial QUality Evaluator (BRISQUE) [6] 2D features extracted from image, 3D features extracted from disparity map distribution, and uncertainty map to train a support vector regression (SVR). Trained SVR detects symmetry or asymmetry of distortion in the left and right images and then assesses the stereo quality. In [39] an NR quality assessment based on Bayesian theory is presented.

It models 3D images using hybrid combination based on posterior and prior feature distributions.

To be able to compare our work with other references, we test our algorithm on two publically available stereo-image databases of LIVE-I [40] and LIVE-II [41]. Image pairs in LIVE-I are symmetrically distorted. Images in LIVE-II are more challenging because asymmetric distortion is present, meaning that one of the two stereo images is distorted while the other one is left intact.

### III. PROPOSED MODEL FOR 3D NR IQA

Here, we use the same assumption as other NR SIQA methods that the stereo images are well calibrated. This means that distortions due to weakly calibrated capturing systems, such as misaligned cameras, are not of issue. Therefore, only the effect of different distortions on binocular and depth perception are to be addressed. As is shown in Fig. 1, in the first stage two synthesized images are formed using the left and right distorted images. The two synthesized images, namely contrast and phase images, are made by a perceptual combination algorithm. In the second stage of our approach, spatial domain natural scene statistics are extracted from each of two binocular combined images. The quality estimation module in Fig. 1 automatically evaluates the quality of the input stereo images. This unit contains three neural networks. Each of the first two ones separately evaluates the feature vectors of one of the synthesized images. The outputs of these networks are fed into a third neural network which produces scores that are very close to human judgments.

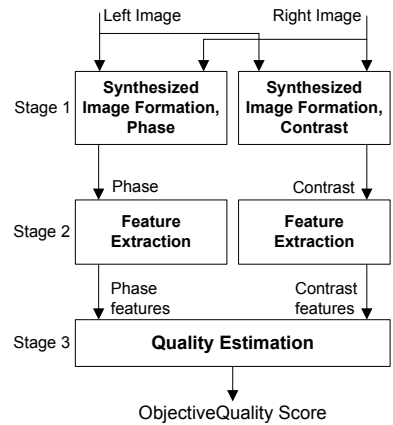

Fig. 1. Block diagram of proposed stereo image quality assessment.

### A. Phase and Contrast Synthesized Images

Here, our goal is to generate a single image (i.e. cyclopean image) from two stereo images. This process, known as binocular single vision, is close to what happens in the brain through sensory fusion and motor fusion when two images are received from the eyes and a single percept is formed. The sensory fusion is performed by the neural elements of the brain and the motor fusion is generated by the correct positioning of the eyes. As the brain eventually combines stereo images, it seems logical to generate a single combined image to evaluate the quality of stereo images. Therefore we propose to perform our evaluations on cyclopean image, generated by a process similar to the motor/sensory fusion process. In [42] a binocular combination algorithm, called DSKL model (Ding, Sperling, Klein, Levi), is presented. They present a model for the process of formation of a single pre-fusion combined image and a model showing what human brain does for fusion of images. We use their post-fusion model to generate our synthesize images. In the followings we briefly explain findings of [42] about how human brain forms a perception.

This model is found by experimenting with human subjects. Two images of sine waves, with different phase shifts, are shown to human subjects. Human perception of these images depends on phase shifts and relative contrast of the images. The left and right images that are shown to human subjects are modeled as:

$$I_L = I_0 + m_L \cos(2\pi f_s x + \theta_L)$$
$$I_R = I_0 + m_R \cos(2\pi f_s x + \theta_R)$$

where $I_0$ is the mean luminance of the sine wave image and $f_s$ is the spatial frequency (in terms of cycles per degree, cpd, with a typical value of 0.68cpd). Also, $m_L$ and $m_R$ are modulation contrasts of the left and right sine waves. Then a simple addition of these two images would generate a single sine wave image of the form:

$$\hat{I} = \widehat{I_0} + \widehat{m} \cos(2\pi f_s x + \hat{\theta})$$

where $\widehat{I_0} = 2I_0$, and

$$\widehat{m} = \sqrt{m_L^2 + m_R^2 + 2m_L m_R \cos(\theta_R - \theta_L)}$$
$$\hat{\theta} = tan^{-1} \frac{m_L \sin\theta_L + m_R \sin\theta_R}{m_L \cos\theta_L + m_R \cos\theta_R}$$

This is a combined image that is formed before sensory/motor fusion. Eyes move and the nervous system adjusts the contrast and a single image is perceived. It is experimentally shown by [42] that the eye movement is a function of disparity $D$ and is controlled by a gain control parameter $\alpha$.

$$\alpha = 1 - \frac{D}{g^2 + D}$$

In this equation $g$ is a threshold at which fusion becomes apparent (typical value of 0.053). If disparity is less than $g^2$ then eye movement and fusion do not occur. When enough disparity exists then fusion occurs and a new combined image, with new phase $\hat{\theta}'$ and contrast $\widehat{m}'$, is generated.

$$\widehat{m}' = \sqrt{m_L^2 + m_R^2 + 2m_L m_R \cos(\alpha(\theta_R - \theta_L))}$$
$$\hat{\theta}' = tan^{-1} \frac{m_L \sin(\alpha\theta_L) + m_R \sin(\alpha\theta_R)}{m_L \cos(\alpha\theta_L) + m_R \cos(\alpha\theta_R)}$$

We consider $\widehat{m}'$ and $\hat{\theta}'$ as two synthetic images. A sample stereo pair, from LIVE-I dataset, is shown in Fig. 2. Distorted versions of this stereo pair are also present in the dataset. In Fig. 3 and Fig. 4 we are showing the contrast and phase images of the stereo pair and its distorted versions. We will statistically show that the effect of a distortion is different in the phase and contrast images. Therefore, contribution of the phase image, for the quality assessment of a stereo image, is different than that of the contrast image.

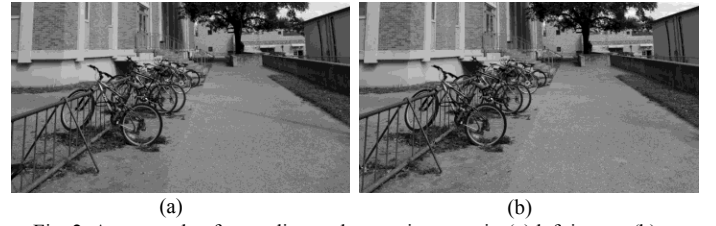

Fig. 2. An example of an undistorted stereo image pair, (a) left image, (b) right image.

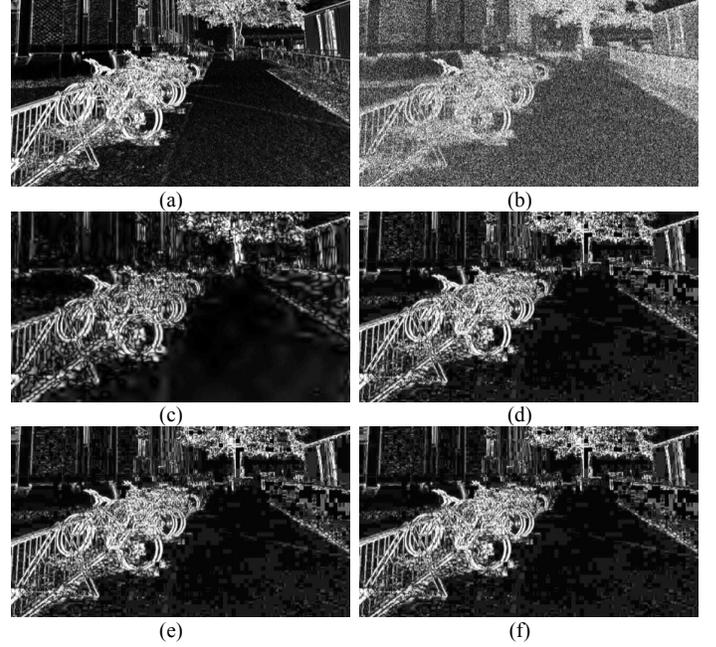

Fig. 3. (a) A sample of generated phase image, and its distorted versions, distorted by: (b) white noise, (c) Jpg2k, (d) Jpg, (e) blur, and (f) fast fading distortion.

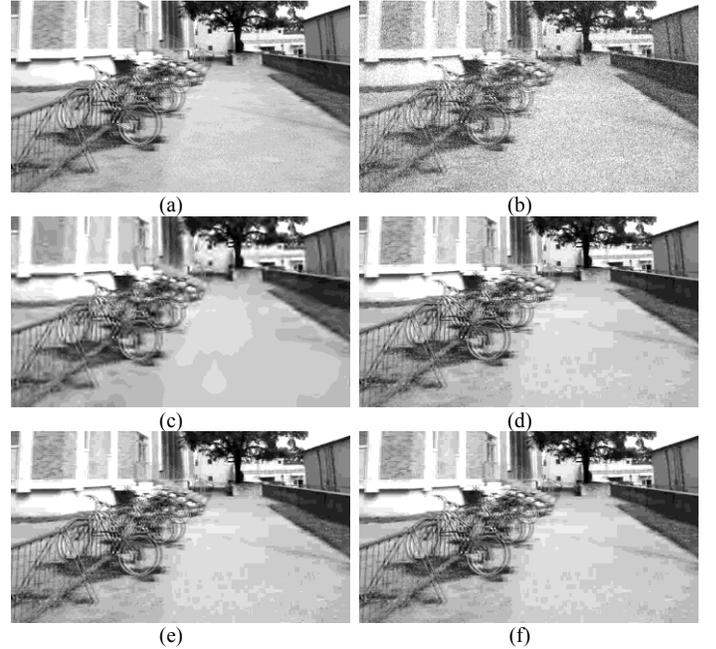

Fig. 4. (a) An example of generated contrast image, and its distorted versions, distorted by: (b) white noise, (c) Jpg2k, (d) Jpg, (e) blur, and (f) fast fading distortion.

## B. Feature Extraction

In our proposed method we study the statistics of the phase and contrast images and based on their behaviors, we fit different distributions to the histograms of the corresponding divisive normalized coefficients. Other distributions are fitted at the next stage on pairwise product of each coefficient with its adjacent coefficients. The characteristic parameters of these distributions are used as two different feature vectors to train a stacked ANN.

### Spatial domain natural scene statistics features

Statistics of natural images follow certain rules [43]. Based on this, various successful 2D NR IQA algorithms have been designed [4-6]. To reduce the correlation between neighboring coefficients of phase, and neighboring coefficients of contrast, a normalization procedure is performed on these coefficients. Hence, mean subtracted contrast normalized (MSCN) coefficients are generated based on the following equation [6]:

$$\text{MSCN}(i,j) = \frac{I(i,j) - \mu(i,j)}{\delta(i,j) + C} \qquad j \in 1\,2 \dots N, i \in 1 \dots M$$

where $I(i,j)$ is the phase/contrast value. Also, $\mu(i,j)$ and $\delta(i,j)$ are representatives of local Gaussian weighted mean and variance values and local variances, respectively. To avoid division by zero, constant $C$ is added to the denominator. The value of $\mu(i,j)$ inside a window is obtained by adding pixels of the window with weights assigned by a 2D circularly-symmetric Gaussian weights.

We observed that MSCN histograms of contrast images have symmetric shapes with zero mean, while those of phase images have asymmetrical shapes and nonzero means. Also the shapes and variances of distributions are different for different distortions depending on the perceptual severity of the distortion. Figure 5 shows normalized histograms of MSCN coefficients for phase and contrast of a sample image. These are for cases when the image is distorted by white noise (WN), JPEG2000 compression (JP2K), JPEG compression (JPEG), Gaussian blur (Blur) and fast-fading (FF).

To model the distribution of MSCN coefficients of contrast, a Generalized Gaussian Distribution (GGD) function, which is symmetric and has a zero mean, is utilized based on the following equation:

$$f(x; \alpha, \sigma^2) = \frac{\alpha}{2\beta\Gamma(1/\alpha)} \exp(-(|\tfrac{x}{\beta}|)^\alpha), \quad \beta = \sigma\sqrt{\frac{\Gamma(\tfrac{1}{\alpha})}{\Gamma(\tfrac{3}{\alpha})}}$$

The parameter $\alpha$ controls the shape of the distribution, $\sigma^2$ represents variance of the distribution, and $\Gamma(\cdot)$ is gamma function, $\Gamma(x) = \int_{x=0}^{\infty} y^{x-1}e^{-y}dy$. For each contrast image the fitted GGD shape parameter, $\alpha$, and variance, $\sigma^2$, are selected as the first two features.

Unlike the contrast, to which symmetric GGD can be fitted, to estimate the shape of an MSCN histogram of the phase images we use an Asymmetric GGD (AGGD). Modeling of AGGD is done based on the following equations:

$$f(x; v, \sigma_l{}^2, \sigma_r{}^2) = \begin{cases} \dfrac{v}{(\beta_l + \beta_r)\Gamma\left(\tfrac{1}{v}\right)} \exp\left(-\left(\dfrac{-x}{\beta_l}\right)^v\right) & x < 0 \\[3mm] \dfrac{v}{(\beta_l + \beta_r)\Gamma\left(\tfrac{1}{v}\right)} \exp\left(-\left(\dfrac{x}{\beta_r}\right)^v\right) & x > 0 \end{cases}$$

$$\beta_l = \sigma_l \sqrt{\frac{\Gamma\left(\tfrac{1}{v}\right)}{\Gamma\left(\tfrac{3}{v}\right)}}, \qquad \beta_r = \sigma_r \sqrt{\frac{\Gamma\left(\tfrac{1}{v}\right)}{\Gamma\left(\tfrac{3}{v}\right)}}, \quad \eta = (\beta_l - \beta_r)\frac{\Gamma\left(\tfrac{2}{v}\right)}{\Gamma\left(\tfrac{1}{v}\right)}$$

In the above equations $v$ is the shape parameter and controls the shape of the distribution, $\sigma_l{}^2, \sigma_r{}^2$ are the distribution variances of left and right respectively. The parameters $\eta, \sigma_l, \sigma_r, v$ are extracted as the first four features of the phase image.

In Fig. 6, we show the ability of these features to separate the distortion in each of the two synthesized images. To generate Fig. 6, we calculated the distributions shown in Fig. 5 and fitted AGGD on the MSCN histograms of phase images and then fitted GGD on the histograms of the contrast images.

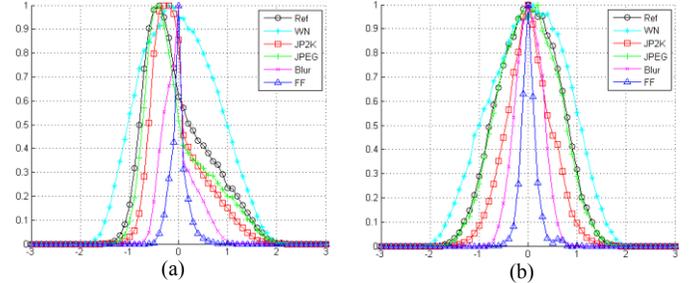

Fig. 5. Normalized histograms of MSCN coefficients for a natural undistorted stereo pair and its various distorted versions, extracted from: (a) phase, and (b) contrast.

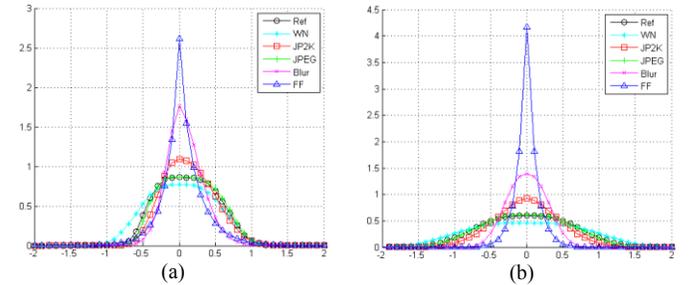

Fig. 6. For a natural undistorted image and its various distorted versions, (a) AGGD fitted to histograms of MSCN coefficients of phase, and (b) GGD fitted to histograms of MSCN coefficients of contrast.

In addition to MSCN coefficients, the statistical relations among neighboring coefficients are also modeled. To this end, an AGGD is fitted to the histogram of the pairwise multiplication of MSCN coefficients in four directions as defined in [6]. We extract a feature vector of 40 elements from the phase image and a vector of 36 features from the contrast image. These features are comprised of distribution shape, mean, and the left/right variances of each four directions, in addition to the AGGD parameters for the phase and GGD parameters for the contrast in two scales. For each of the two synthesized images of phase and contrast the normalized histograms, of pairwise multiplications in one of the four directions, are shown for different distortions in Fig. 7. The corresponding fitted AGGDs are exhibited in Fig. 8.

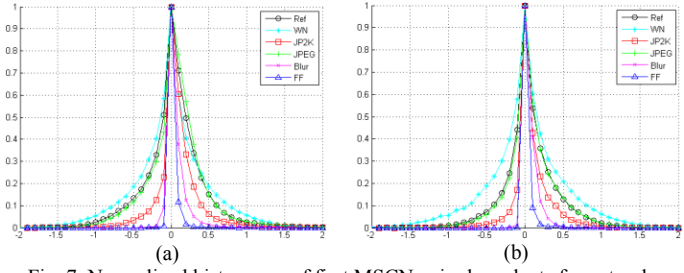

Fig. 7. Normalized histograms of first MSCN paired-products for natural undistorted images and their various distorted versions, extracted from: (a) phase, and (b) contrast.

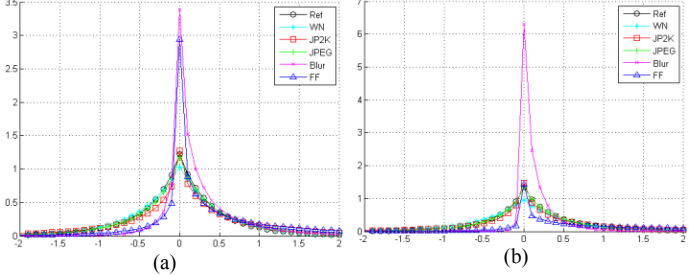

Fig. 8. AGGD fitted to histograms of first MSCN *paired-products* for natural undistorted image and its various distorted versions, extracted from: (a) phase image, and (b) contrast image.

Images that are used in these experiments are from LIVE-I and LIVE-II datasets. To determine the effect of a feature on the process of stereo quality assessment, the correlation is measured between each feature and subjective scores, in all images. If a feature has a high correlation with the Mean Opinion Scores (MOS), then the presence of such feature in a vector will result in objective quality scores that are close to subjective ones. Figure 9 shows the Pearson's Linear Correlation Coefficient (PLCC) values of the features of phase and contrast, with subjective quality scores for images with different distortion types. Parts (a) and (b) show correlation values of phase and contrast features respectively, for images of LIVE-I database.

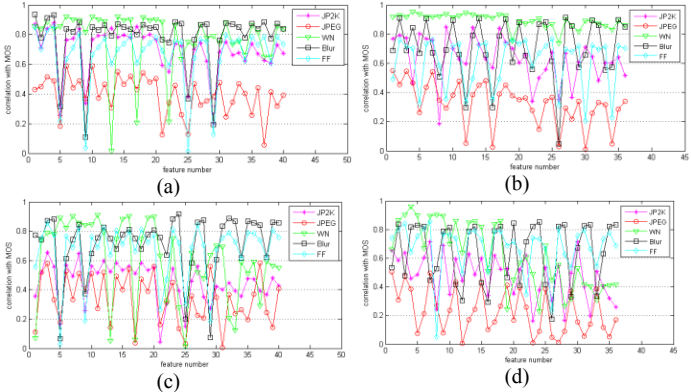

Fig. 9. PLCC plots. Correlations of MOS with features extracted from (a) phase, and (b) contrast, calculated for LIVE-I distorted-image dataset.
Correlations of MOS with features extracted from (c) phase, and (d) contrast, calculated for LIVE-II distorted-image dataset.

Likewise, parts (c) and (d) of Fig. 9 show correlations for the LIVE-II database. High correlation levels in these graphs indicate that, for both LIVE-I and LIVE-II databases [40, 41], features obtained from synthesized images are highly effective. The collection of these features is a good descriptor of 3D perceptual quality of stereo images. The extracted features, from images with white noise, have the highest correlation with visual quality features. The lowest correlation is for the features obtained from JPEG compressed images.

### C. Quality estimation

Stacked generalization, which was first presented by Wolpert [44], is an effective method for combining a number of generalizers. It uses partitioning of training data or the feature space to elevate the generalization performance of the whole system. It consists of a number of "level-0" generalizers where each, is independently trained with a subset of the available features. Outputs of level-0 generalizers are combined by going into "level-1". This combination of level-0 outputs is not just a linear combination but it is a means of combining a number of generalizers to produce a new one. Partitioning of the feature space at the level-0 should be in a way that a complex computation task is partitioned into a number of computationally simple tasks. Hence, by combination of the results of level-0 generalizers the desired solution of the initial task is obtained.

Since ANNs are powerful tools for nonlinear approximations and they mimic real-time complex biological human decision system, they are good candidates for stacked generalization. We propose to use three feed-forward ANNs, two at level-0 and one at level-1.

As shown in Fig. 10, each of the two level-0 parallel ANNs receives features of one of the synthesized images. The first neural network gets the features of the phase, and the second one is fed by the features of the contrast image. Each of these networks is trained with train set images. The third network, the refiner in level-1, is trained with the test results of the first two networks. In other words, the third network, as an expert quality assessor, learns how to correct the opinions of other experts to achieve better results.

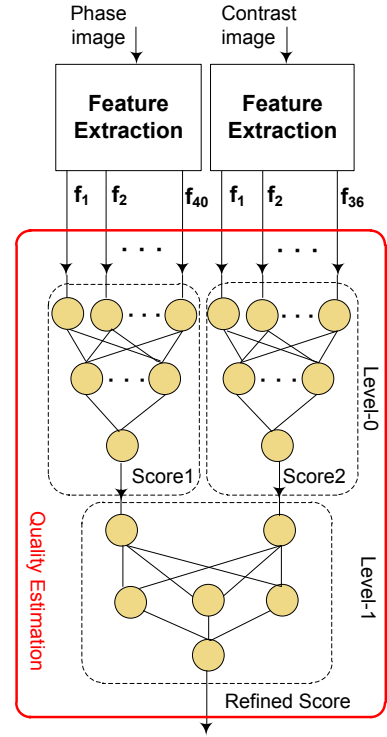

Fig. 10. Quality estimation stage based on stacked neural network structure.

## IV. EXPERIMENTAL RESULTS

Here we evaluate the performance of our NR SIQA proposed method. First, each of the two level-0 ANNs independently, and once all of the three ANNs in stacked model together, are trained and tested.. Quality scores obtained by the proposed method have been compared with the best results obtained from the best assessment methods in this area.

### A. Databases

In the following we explain characteristics of LIVE-I and LIVE-II databases.

#### 1) LIVE-I 3D Image Quality Database

LIVE-I 3D database contains 20 reference and 365 symmetrical distorted stereo images. Among these distorted images, 80 pairs are allocated to the following distortions: JPEG compression (JPEG), JPEG2000 (JP2K) compression, additive White Gaussian Noise (WN), and Fast-Fading (FF) model based on the Rayleigh fading channel. Also, 45 pairs are dedicated to Gaussian Blur (Blur) with different distortion levels. All of the damaged stereo images in this database are symmetrically distorted. It means that both the left and right images have been equally affected by the same distortion process. In addition, the subjective quality scores of difference MOS are in the range of -10 to 60.

#### 2) LIVE-II 3D Image Quality Database

The LIVE-II 3D database contains 8 reference and 360 distorted stereo images. In the database, for each type of distortion, in addition to three pairs of symmetric distorted versions, there are six pairs of asymmetric distorted ones. This means that left and right images are distorted at different levels or one of them has full quality while the other's quality is degraded. This characteristic makes the quality assessment of the stereo images in the LIVE-II dataset more challenging. Fig. 11 shows samples of symmetric and asymmetric distorted images from the LIVE-I and LIVE-II datasets. The human based DMOS scores are in the range of 0 to 100.

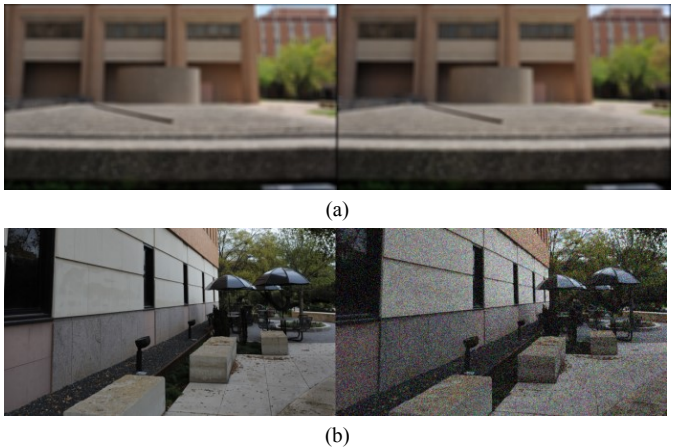

(a)

(b)

Fig. 11. (a) Symmetrically distorted image-pair from LIVE-I dataset (Blur).
(b) Asymmetrically distorted image-pair from LIVE-II dataset (WN).

### B. Network training and test

Each of the single models includes a two layer feed forward neural network, with sigmoid activation function neurons in its hidden layer, and linear output layer. Each neural network is configured for a particular application through the process of training. Training will continue until the error subsides to a desired value. To avoid overtraining of networks, we used Resilient back Propagation (RP) training law. The extracted features from the two synthesized images of phase and contrast are utilized separately by the two level-0 networks with the same number of 25 neurons in the hidden layer. The number of neurons for the level-1 refiner network is also set to three.

Our no-reference method requires training to determine the optimum weights for the two layers of ANNs. To evaluate each of the single models, like most of the learning based IQA methods [6, 9, 38, 39] we train each one using 80 percent of database images, which are randomly selected. The remaining 20 percent of images are used as the test set. There is no overlap between the selected training images and those used for testing. To ensure independence of the results from the set of images selected for training or test, the train-test process was repeated 1000 times. Median of all obtained results is reported as the final result [6, 9, 38, 39]. The entire operation for each model is independently done on the LIVE-I and LIVE-II databases. The results for each data set will later be compared with other assessment methods.

### C. Performance Evaluation of Model

We compare the power of the models using only the features of phase, contrast and also the influence of the third corrector ANN to improve the assessment efficiency. Scatter plots of objective scores versus subjective scores in a single train-test process, for the first two single models and also the final model on LIVE-I and LIVE-II datasets are displayed in Fig. 12.

In plots of Fig. 12, the vertical axis denotes the subjective ratings of the perceived distortions and the horizontal one indicates the corresponding predicted quality scores. If the subjective and objective scores are exactly equal, then all points will be on the $y = x$ line (i.e. red dashed line). Hence, scattering of points close to the bisector of the first quadrant is an indication of better performance of that approach. The RMSE, SRCC, PLCC and the equation of the best fitted (the black solid) line to the data are calculated in each plot. As can be seen, the results of the stacked model, consisting of the stacked ANN, are more correlated with the visual scores as compared to the initial simple models. In most cases, the fitted line almost matches the bisector line that shows the ability of the model to predict correct image quality scores. Moreover, lower RMSE values in both LIVE-I and LIVE-II databases are another evidence of this claim.

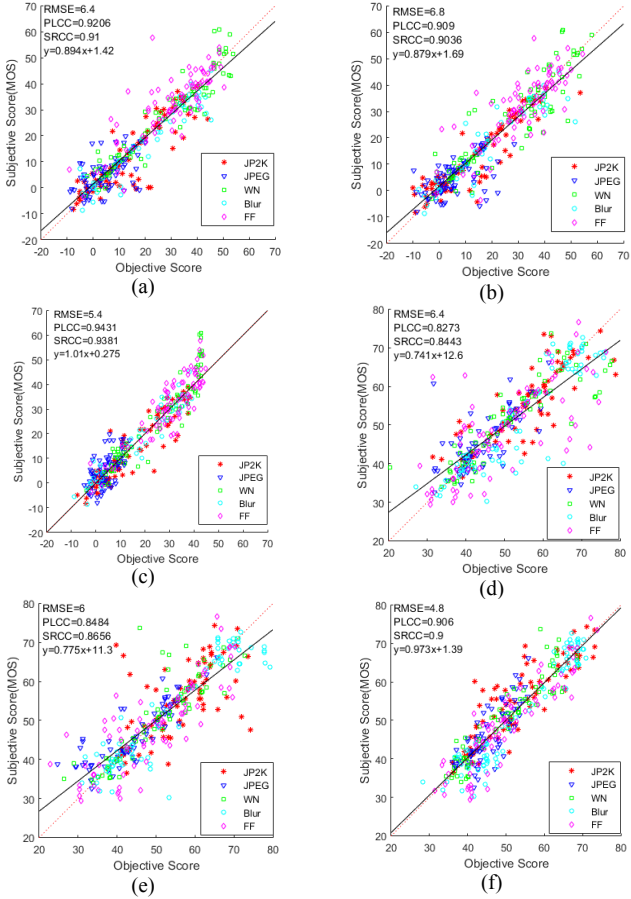

Fig. 12. Scatter plots of objective scores versus subjective scores for (a) phase model, (b) contrast model, (c) stacked model on LIVE-I dataset, (d) phase model, (e) contrast model, (f) stacked model on LIVE-II dataset.

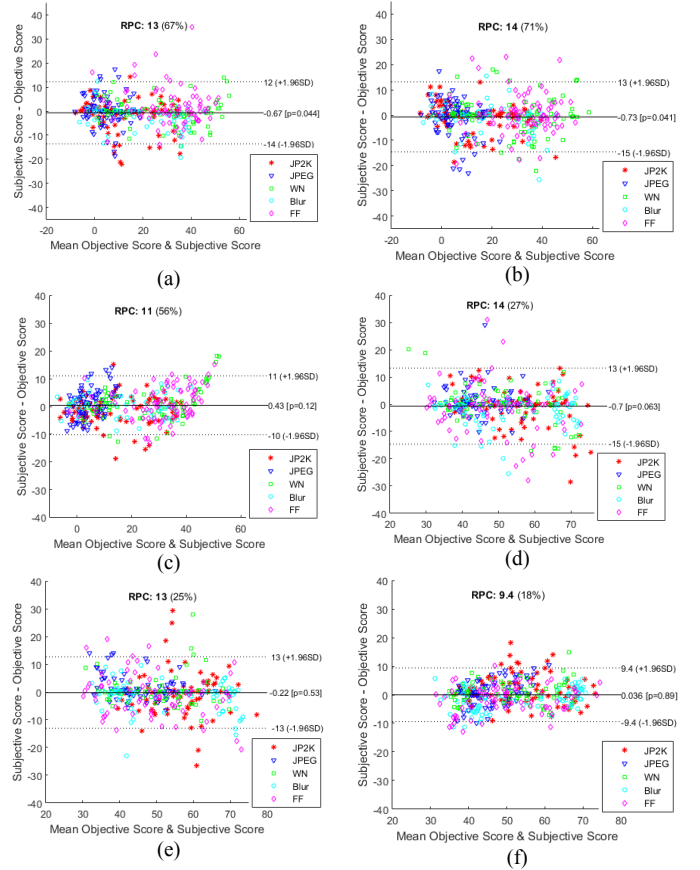

Fig. 13. Bland-Altman plots of (a) phase model, (b) contrast model, (c) stacked model on LIVE-I dataset, (d) phase model, (e) contrast model, (f) stacked model on LIVE-II dataset.

The Bland-Altman [45] plots of the same data are shown in Fig. 13. In this graphical model the differences between subjective and objective scores are plotted against the average values of these scores. The horizontal solid line indicates the mean value of the difference between subjective and objective scores. A distance of 1.96 times the standard deviation of difference values is called Re-Producibility Coefficient (RPC). The two dashed lines, called the limits of agreement, are drawn at a distance of plus and minus RPC from the mean difference line. Narrower RPC values show greater capability of stacked model for quality assessment of stereo images.

### D. Comparison to 2D State of the Art Methods

In this section we compare our proposed model with the state of the art 2D IQA methods. PSNR, SSIM [2] and MS-SSIM [3] are full reference 2D IQA metrics. We also compare our method with three no-reference methods of BRISQUE (BLIND/Referenceless Image Spatial Quality Evaluator [6]), DIIVINE (Distortion Identification-based Image Verity and Integrity Evaluation [4]), and BLIINDS-II (Blind Integrity Notator using DCT Statistics-II [5]). In order to apply these approaches to 3D images, we separately estimate the quality of left and right images and the average of them is reported as the 3D quality of stereo pair. Again, we want to find the correlation between the features that these methods offer with subjective scores. For example, we want to know how correlated PSNR values of images are with subjective scores.

Hence, the overall PLCC and SROCC are calculated, as shown in Table 4. We tested our stacked method and the mentioned FR and NR features on the images of the LIVE-I and LIVE-II databases. The accuracy of our prediction is higher than other 2D methods. The superiority is more visible for the images of the LIVE-II dataset. This indicates that only averaging the 2D qualities is not adequate to describe the quality of stereo images especially for the asymmetrically distorted stereo images. Our proposed features extracted from the phase and contrast images include a wide range of 2D and 3D aspects of the binocular perception that result in high performances for both symmetric and asymmetric distorted stereo pairs.

Table 4. PLCC and SROCC values of our models and state of the art 2D IQA methods, performed on LIVE-I and LIVE-II databases.

| Database | | LIVE-I | | LIVE-II | |
|---|---|---|---|---|---|
| Type | Method | PLCC | SROCC | PLCC | SROCC |
| FR | PSNR | 0.834 | 0.834 | 0.665 | 0.665 |
| | SSIM [2] | 0.872 | 0.876 | 0.792 | 0.792 |
| | 2D MS-SSIM [3] | 0.926 | 0.926 | 0.777 | 0.776 |
| NR | BRISQUE [6] | 0.910 | 0.901 | 0.749 | 0.701 |
| | DIIVINE [4] | 0.939 | 0.929 | 0.697 | 0.669 |
| | BLIINDS-II [5] | 0.917 | 0.910 | 0.736 | 0.700 |
| | Stacked Model | **0.955** | **0.945** | **0.923** | **0.913** |

*E. Comparison to State of the Art 3D Methods*

Here, we compare our NR method with four full referenced methods (FR) [17, 25, 26, 29], two reduced reference methods (RR) [32, 33] and three no-reference methods (NR) [9, 38, 39] for stereo image quality assessment. The first FR method is based on SSIM and additional depth information [17], the second is the cyclopean MS-SSIM [25] which considers the effect of binocular rivalry on 3D quality of stereo images. The third algorithm is the SDM-GSSIM [26]. The forth algorithm is based on the difference of sparse coefficient vectors of the reference and distorted pairs proposed in [29]. The RR method proposed in [32] uses the edge information of depth and color data in the corresponding areas of images. Algorithm of [33] utilizes parameters that are extracted from the image and its depth in the contourlet domain. Among the existing no-reference techniques, best results belong to an application specific algorithm proposed by Akhter [38], and two general purpose algorithms proposed by Chen [9] and Shao [39]. Similar to most quality assessment techniques, the Spearman's Rank Ordered Correlation Coefficient (SROCC), Pearson's Linear Correlation Coefficient (PLCC) and Root Mean Squared Error (RMSE) are used to evaluate the performance. Closer values of PLCC and SROCC to 1 indicate higher correlations to subjective values. On the other hand, smaller quantities of RMSE are more desired. Our results of PLCC, SROCC and RMSE are compared with the above mentioned methods, reported in [9, 39, 26, 29, 33].

*1) Performance Comparison on LIVE-I Database*

PLCC, SROCC and RMSE results of our models, tested on LIVE-I dataset, are evaluated and compared with other methods in the left parts of Tables 1, 2, and 3 respectively. The best two NR/RR results in each column are bolded and the best FR one is marked in italic.

It can be seen that if we use one of the synthesized images, the results would be comparable with the referential and non-referential methods. The results that we obtain by our final stacked model, using both synthesized images, are better than the first two single models and outperform even the best FR results for all distortions. In comparison with the state of the art NR SIQA methods, our method defeats all methods in WN and FF distortions. It also produces results that are very close to [39] for the other three distortions. The overall results, in the last column, indicate that in average, our final model has at least 2% better PLCC, SROCC results and at least one unit lower RMSE values than FR methods.

*2) Performance Comparison on LIVE-II Database*

Similar to LIVE-I dataset, PLCC, SROCC and RMSE results are measured, after we applied our models to the images of the LIVE-II dataset. The right half of Tables 1, 2, and 3 show our results as compared to other assessment algorithms. This dataset has added asymmetrically distorted images, which has made their quality assessment much more complicated. Our two initial models, which use only one synthesized image, perform below that of Chen's [9] and the cyclopean MS-SSIM [25]. But our main stacked model produces better results than the mentioned references. In the group of NR/RR methods, our results are always one of the two best results for all distortions. Our stacked model surpasses the best existing PLCC and SROCC values by more than 2%. Our overall results are also superior to all other methods by at least 3%. However, FR methods, which have access to the reference image, are expected to have better performance than the NR and RR techniques. Nevertheless, our NR approach not only has better results than NR methods, but it also defeats all FR methods in each distortion. This is due to the use of proper descriptors, strong stacked ANN structure, and the use of appropriately formed synthesized images. These synthesized images reveal the effects of different distortions on the perceived image.

## V. CONCLUSIONS

We proposed a no-reference stereo image quality assessment method. Inspired by the sensory-motor fusion of the brain, we performed binocular combination of every stereo-image pair to produce a pair of synthesized images (phase and contrast). These images proved to be highly valuable in exposing the type and severity of different distortions. We analyzed the spatial domain statistics of the generated phase and contrast images. Two different feature vectors can be extracted from every two synthetic images. These features are sensitive to the changes caused by different distortions. We showed that these 2D sets of features, extracted from the phase and contrast images, can provide highly reliable 3D quality assessment measures. The proposed quality assessment was implemented by two parallel ANN channels as the first layer of the proposed quality estimation structure. The resulting scores were refined using a second layer. The results showed that our model significantly outperforms most of the state of the art 3D image quality assessment methods including no-reference, reduced-reference and full-reference ones. The synthesized images used in this method were generated simply and quickly. The contrast and phase images were synthesized in 0.37s by a 3.4 GHz, Core i7 computer with 16 GB RAM. Whereas, the average production time for a Gabor cyclopean used in [9] was 20.72s on the same hardware platform. In addition, our approach has the potential of parallel implementation. This is true for the formation of the synthesized images, as well as the feature extraction part of the algorithm. Our method basically uses spatial domain features, which have lower computational complexity than the transform domain features. Furthermore, direct quality assessment is performed without the need for an initial classification of the distortion type and/or the asymmetry of the distortion.

Table 1. PLCC values of our models and other 3D IQA methods, performed on LIVE-I and LIVE-II databases.

| Database | | LIVE-I | | | | | | LIVE-II | | | | | |
|---|---|---|---|---|---|---|---|---|---|---|---|---|---|
| Type | Method | WN | JP2K | JPEG | Blur | FF | All | WN | JP2K | JPEG | Blur | FF | All |
| FR | 3D-MS-SSIM [25] | 0.942 | 0.912 | 0.603 | 0.942 | 0.776 | 0.917 | *0.957* | *0.834* | *0.862* | *0.963* | 0.901 | *0.900* |
| | Benoit [17] | 0.925 | 0.935 | 0.640 | 0.948 | 0.747 | 0.902 | 0.926 | 0.784 | 0.853 | 0.535 | 0.807 | 0.748 |
| | SDM-GSSIM [26] | 0.935 | *0.940* | *0.671* | 0.952 | *0.865* | 0.933 | - | - | - | - | - | - |
| | FR-Shao[29] | *0.945* | 0.921 | 0.520 | *0.959* | 0.859 | *0.935* | 0.946 | 0.782 | 0.747 | 0.958 | *0.905* | 0.863 |
| RR | Hewage [32] | 0.895 | 0.904 | 0.530 | 0.798 | 0.669 | 0.830 | 0.891 | 0.664 | 0.734 | 0.450 | 0.746 | 0.558 |
| | Wang [33] | 0.913 | 0.916 | 0.570 | 0.957 | 0.783 | 0.892 | - | - | - | - | - | - |
| NR | Akhter [38] | 0.904 | 0.905 | 0.729 | 0.617 | 0.503 | 0.626 | 0.722 | 0.776 | 0.786 | 0.795 | 0.674 | 0.568 |
| | Chen [9] | 0.917 | 0.907 | 0.695 | 0.917 | 0.735 | 0.895 | 0.947 | **0.899** | **0.901** | **0.941** | **0.932** | **0.895** |
| | NR-Shao [39] | 0.938 | **0.950** | 0.796 | **0.986** | 0.837 | 0.957 | - | - | - | - | - | - |
| | Phase Model | 0.935 | 0.887 | 0.710 | 0.924 | 0.829 | 0.918 | 0.924 | 0.800 | 0.763 | 0.917 | 0.831 | 0.854 |
| | Contrast Model | **0.945** | 0.907 | **0.772** | 0.927 | 0.834 | 0.933 | **0.948** | 0.820 | 0.788 | 0.913 | 0.839 | 0.869 |
| | Stacked Model | **0.955** | **0.939** | 0.771 | **0.959** | **0.882** | 0.956 | **0.966** | **0.897** | **0.866** | **0.957** | **0.918** | **0.923** |

Table 2. SROCC values from our models and other 3D IQA methods, performed on LIVE-I and LIVE-II databases.

| Database | | LIVE-I | | | | | | LIVE-II | | | | | |
|---|---|---|---|---|---|---|---|---|---|---|---|---|---|
| Type | Method | WN | JP2K | JPEG | Blur | FF | All | WN | JP2K | JPEG | Blur | FF | All |
| FR | 3D-MS-SSIM [25] | *0.948* | 0.888 | 0.530 | 0.925 | 0.707 | 0.916 | 0.940 | *0.814* | 0.843 | 0.908 | 0.884 | *0.889* |
| | Benoit [17] | 0.930 | 0.910 | *0.603* | 0.931 | 0.699 | 0.899 | 0.923 | 0.751 | *0.867* | 0.900 | *0.933* | 0.880 |
| | SDM-GSSIM [26] | - | - | - | - | - | *0.925* | - | - | - | - | - | - |
| | FR-Shao[29] | 0.941 | *0.894* | 0.495 | *0.940* | *0.796* | 0.903 | *0.965* | 0.785 | 0.733 | *0.920* | 0.891 | 0.849 |
| RR | Hewage [32] | **0.940** | 0.856 | 0.500 | 0.690 | 0.545 | 0.814 | 0.880 | 0.598 | 0.736 | 0.028 | 0.684 | 0.501 |
| | Wang [33] | 0.907 | 0.883 | 0.542 | 0.925 | 0.655 | 0.889 | - | - | - | - | - | - |
| NR | Akhter [38] | 0.914 | 0.866 | 0.675 | 0.555 | 0.640 | 0.383 | 0.714 | 0.724 | 0.649 | 0.682 | 0.559 | 0.543 |
| | Chen [9] | 0.919 | 0.863 | 0.617 | 0.878 | 0.652 | 0.891 | **0.950** | **0.867** | **0.867** | **0.900** | **0.933** | **0.880** |
| | NR-Shao [39] | 0.935 | **0.936** | **0.818** | **0.927** | **0.814** | **0.950** | - | - | - | - | - | - |
| | Phase Model | 0.928 | 0.868 | 0.694 | 0.882 | 0.780 | 0.927 | 0.919 | 0.790 | 0.743 | 0.840 | 0.816 | 0.847 |
| | Contrast Model | 0.938 | 0.884 | **0.756** | 0.891 | 0.775 | 0.927 | 0.936 | 0.809 | 0.740 | 0.818 | 0.831 | 0.865 |
| | Stacked Model | **0.945** | **0.915** | 0.750 | **0.919** | 0.837 | **0.947** | **0.953** | **0.875** | **0.832** | **0.874** | **0.907** | **0.913** |

Table 3. RMSE values from our models and other 3D IQA methods, performed on LIVE-I and LIVE-II databases.

| Database | | LIVE-I | | | | | | LIVE-II | | | | | |
|---|---|---|---|---|---|---|---|---|---|---|---|---|---|
| Type | Method | WN | JP2K | JPEG | Blur | FF | All | WN | JP2K | JPEG | Blur | FF | All |
| FR | 3D-MS-SSIM [25] | *5.581* | 5.320 | 5.216 | 4.822 | *7.837* | 6.533 | *3.368* | *5.562* | *3.365* | *3.747* | *4.966* | *4.987* |
| | Benoit [17] | 6.307 | *4.426* | *5.022* | *4.571* | 8.257 | 7.061 | 4.028 | 6.096 | 3.878 | 11.763 | 6.894 | 7.490 |
| | SDM-GSSIM [26] | 7.853 | 5.909 | 6.465 | 5.919 | 8.312 | 7.857 | - | - | - | - | - | - |
| | FR-Shao[29] | - | - | - | - | - | *5.816* | - | - | - | - | - | 5.706 |
| RR | Hewage [32] | 7.405 | 5.530 | 5.543 | 8.748 | 9.226 | 9.139 | 10.713 | 7.343 | 4.976 | 12.436 | 7.667 | 9.365 |
| | Wang [33] | 6.777 | **5.189** | 5.374 | **4.178** | 7.725 | 7.408 | - | - | - | - | - | - |
| NR | Akhter [38] | 7.092 | 5.483 | **4.273** | 11.387 | 9.332 | 14.827 | 7.416 | 6.189 | 4.535 | 8.450 | 8.505 | 9.294 |
| | Chen [9] | 6.433 | 5.402 | 4.523 | 5.898 | 8.322 | 7.247 | **3.513** | 4.298 | 3.342 | **4.725** | **4.180** | **5.102** |
| | NR-Shao [39] | - | - | - | - | - | - | - | - | - | - | - | - |
| | Phase Model | 6.199 | 6.799 | 4.902 | 6.488 | 7.832 | 6.686 | 4.482 | 6.867 | 5.433 | 6.043 | 6.860 | 6.282 |
| | Contrast Model | **5.692** | 6.006 | 4.404 | 5.809 | **7.237** | **6.685** | 3.638 | 6.247 | 5.070 | 6.083 | 6.780 | 6.280 |
| | Stacked Model | **5.017** | **4.644** | **4.290** | **4.458** | **5.997** | **4.998** | **2.936** | 5.083 | **4.071** | **4.581** | **4.974** | **4.436** |